\documentclass[acmsmall, sigconf,screen]{acmart}
\renewcommand\footnotetextcopyrightpermission[1]{} 
\AtBeginDocument{%
  \providecommand\BibTeX{{%
    \normalfont B\kern-0.5em{\scshape i\kern-0.25em b}\kern-0.8em\TeX}}}

\usepackage{graphicx} 
\usepackage{pgfplots}
\usepackage{amsmath,amssymb}
\DeclareMathOperator{\E}{\mathbb{E}}
\usepackage{breqn}

\usepackage{amsthm}
\usepackage{hyperref}
\usepackage[utf8]{inputenc}
\usepackage[english]{babel}
\usepackage{mathtools}
\newcounter{l2}
\newcommand{\balphlist}{\begin{list}{(\alph{l2})}{\usecounter{l2}}}
\newcommand{\barablist}{\begin{list}{\arabic{l1}}{\usecounter{theorem}}}
\newcounter{foo}
\newcounter{theorem}

\usepackage{setspace} 
\usepackage{array} 
\usepackage{paralist} 
\usepackage{verbatim} 
\usepackage{subfig} 
\usepackage{amsthm} 
\usepackage{amssymb}
\usepackage{mathtools}
\usepackage{todonotes}

\usepackage{bbm}
\usepackage{fancyhdr} 
\usepackage[nottoc,notlof,notlot]{tocbibind} 
\usepackage[titles,subfigure]{tocloft} 

\usepackage{ marvosym }

\theoremstyle{definition}

\copyrightyear{2021}
\acmYear{2021}
\setcopyright{rightsretained}

\DeclareUnicodeCharacter{2212}{-}

\begin{document}
\title{Using Meta Reinforcement Learning to Bridge the Gap between Simulation and Experiment in Energy Demand Response}

\settopmatter{authorsperrow=1} 
\newcommand{\tsc}[1]{\textsuperscript{#1}} 
\author{Doseok Jang\tsc{1}, Lucas Spangher \tsc{1}, Manan Khattar \tsc{1}, Utkarsha Agwan \tsc{1}, Costas Spanos \tsc{1}} 
\affiliation{
  \institution{ 1. Department of Electrical Engineering and Computer Sciences, University of California, Berkeley}
  \city{Berkeley}
  \state{California}
  \country{USA}
}

\begin{abstract}
Our team is proposing to run a full-scale energy demand response experiment in an office building. Although this is an exciting endeavor which will provide value to the community, collecting training data for the reinforcement learning agent is costly and will be limited. In this work, we apply a meta-learning architecture to warm start the experiment with simulated tasks, to increase sample efficiency. We present results that demonstrate a similar a step up in complexity still corresponds with better learning.  
\end{abstract}

\keywords{prosumer, aggregation, reinforcement learning, microgrid, transactive energy}

\acmConference[e-Energy '21]{The Twelfth ACM International Conference on Future Energy Systems}{June 28-July 2, 2021}{Virtual Event, Italy}
\acmBooktitle{The Twelfth ACM International Conference on Future Energy Systems (e-Energy '21), June 28-July 2, 2021, Virtual Event, Italy}\acmDOI{10.1145/3447555.3466589}
\acmISBN{978-1-4503-8333-2/21/06}

\maketitle

\pagestyle{plain}

 \section{Introduction} \label{sec:intro}
 
The bridge from simulation to experiment can be difficult to cross for a variety of reason, both statistical and practical. When considering experiments of phenomena in the energy grid, techniques to help a learned controller retain some information gained during simulation can be very beneficial for reducing these practical considerations.  

We focus here on the electrical grid. As the grid decarbonizes, volatile resources like wind and solar will replace on-demand resources like fossil fuels, and there arises a mismatch between generation and demand. Grids that do not adequately prepare for this question will face daunting consequences, from curtailment of resources \citep{spangher2020prospective} to voltage instability and physical damage, despite having adequate generative capability. Indeed, these problems will only grow larger and demand more solutions as the world moves away from fossil fuels.  
 
 Demand response, a strategy in which customers are incentivized to shift their demand for energy resources to parts of the day where generation is plentiful, is seen as a common solution to the problem of generation volatility. Given the lack of needed material infrastructure and cheapness of the incentives, it has several positives above physical energy storage systems. 
 
 Building energy is a primary target of demand response, and both the central administration of signals and building-level response has been thoroughly studied in residential and industrial settings (\citep{asadinejad2018evaluation}, \citep{ma2015cooperative}, \citep{li2018integrating}, \citep{yoon2014dynamic}, \citep{johnson2015dynamic}.) However, while physical infrastructures\citep{das2020occupants} of office buildings have been studied for demand response (\citep{8248801}), there has been no large scale experiment aimed at eliciting a behavioral demand response. The lack of an office-centered study is understandable when we consider that most offices do not have a mechanism to pass energy prices onto workers. If they did, however, not only would a fleet of decentralized batteries -- laptops, cell phone chargers, etc. be able to be coordinated to function as a large deferable resource, but building managers could save money by adapting their buildings' energy usage to a dynamic utility price.

The SinBerBEST collaboration has developed a Social Game\citep{konstantakopoulos2019design} that facilitates workers to engage in a competition around energy \citep{konstantakopoulos2019deep}, \citep{das2019novel}. Through this framework, a first-of-its-kind experiment has been proposed to implement behavioral demand response within an office building \citep{spangher2020prospective}. Prior work has proposed to describe an hourly price-setting controller that learns how to optimize its prices \citep{spangher2020augmenting}. However, given the costliness of iterations in this experiment and the work that has been put into building a complex simulation environment, warm-starting the experiment's controller with learning from the simulation could prove valuable to its success.  

We report a simulated warm-started experiment. We will in Section \ref{sec:background} contextualize the architecture of our reinforcement learning controller within reinforcement learning and meta-learning. In Section \ref{sec:methods} we will describe the simulation setup and the way we test the warm-started controller. In Section \ref{sec:results} we will give results. Finally, in Section \ref{sec:Discussion} we will discuss implications of the controller and the future work this entails. 
 
 \section{Background} \label{sec:background}

 \subsection{Reinforcement Learning}
 
  Reinforcement learning (RL) is a type of agent-based machine learning where control of a complex system requires actions that optimize the system \citep{sutton2018reinforcement}, i.e. they seek to optimize the expected sum of rewards for actions ($a_t$) and states ($s_t$) in a policy parameterized by $\theta$; i.e.,  $J(\theta) = \E(\sum_{s_t,a_t \sim p_{\pi}}[r(s_t, a_t)])$. Often we model a policy as a deep neural network with weights $\theta$ that takes states as input and outputs actions for each state. RL is useful in contexts where actions and environments are simple or data is plentiful, with early use cases being optimizing the control of backgammon \citep{tesauro1994td}, the cart-pole problem, and Atari \citep{mnih2013playing}. Recently, RL has shown promise in solving much more difficult problems like optimizing the control of Go beyond human capability \citep{silver2016mastering}. In addition, much work has been done trying to extend RL to other situations. 
  
 Policy gradient methods are a class of RL algorithms used to train policy networks that suggest actions (\emph{actors}) via critic networks, trust regions, and other methods. A variant of these methods, Proximal Policy Optimization (PPO) algorithms (\cite{schulman2017proximal}), optimizes a surrogate loss 
 \begin{equation}
    L_{CLIP}(\theta)=E_t[min(r_t(\theta) \hat{A}_t, clip(r_t(\theta),1 − \epsilon,1 + \epsilon) \hat{A}_t)]
 \end{equation}
 which enables multiple gradient updates to be done on these actors using the same samples. However, even with PPO optimization, several months worth of real-world training data would have to be collected to fully train an hourly price-setting controller \citep{spangher2020augmenting} in our Social Game. We seek to leverage a detailed simulation with some behaviorally reasonable dynamics encoded in a model that can train on both simulation environments and experimental environments. We thus propose the use of Model Agnostic Meta Learning, a few-shot learning algorithm which trains an RL meta-architecture that learns to warm start a model's parameters as it trains over a distribution of similar tasks. We propose a MAML optimization procedure over a PPO agent architecture.  
  
  \subsection{Model Agnostic Meta-Learning (MAML)}

  Model Agnostic Meta-Learning (MAML) (\cite{finn2017model}) is a meta-learning algorithm that seeks to learn a neural network weight initialization $\theta$ that allows for fast adaptation to different types of tasks. We first define a task as a transition distribution $q_i(x_{t+1}|x_t, a_t)$ and a reward function $R_i$. MAML can be implemented over any neural network model. In the RL variant, given a set of tasks $\tau$ and a distribution of those tasks $p(\tau)$ MAML seeks to find an initialization $\theta$ that can achieve as high a reward as possible after $K$ gradient update steps in any task. To maximize reward after K updates, MAML optimizes a policy network, $f_{\phi}$ for $K$ gradient update steps starting from the learned initialization $\theta$ and computes the following meta loss:
  \begin{equation}
  L_{\tau_i}(f_{\theta}) = - \mathbb{E}_{x_t, a_t \sim f_{\phi}, q_{\tau_i}}[\sum^{H}_{t=1} R_i(x_t, a_t)]
  \end{equation}
  MAML then optimizes $\theta$ by gradient descent on the sum of the meta-losses over all tasks, eventually computing parameters $\theta$ that produce good results after just $K$ update steps.
  Training MAML on a PPO architecture could be described as consisting of an inner adaptation step and an outer, meta adaptation step. In the inner adaptation step, PPO is trained on a task $\tau_i$, randomly chosen according to $p(\tau)$ for $K$ PPO update steps, starting from the weight initialization MAML is optimizing. In the outer adaptation step, the weight initialization is updated according to the training trajectories of PPO that were collected in the inner adaptation steps. These outer and inner adaptation steps are alternated to compute the weight initialization that produce the highest reward after PPO is trained starting from it for $K$ steps, on a task from the task distribution $\tau$. We experiment in this paper in observing the performance of a trained MAML+PPO weight initialization on several tasks that are outside of the task distribution $\tau$ but are still within our simulated Social Game setting.

  \subsection{RL in Demand Response}
  
  RL has been applied to a number of demand response situations, but almost all work centers on agents that directly schedule resources \citep{6963416}, \citep{7018632}, \citep{6848212}, \citep{6915886}, \citep{RAJU2015231}, \citep{FUSELLI2013148}. RL architectures can vary widely, for example Kofinas et. al. deploy a fuzzy Q-learning multi-agent that learns to coordinate appliances to increase reliability \citep{KOFINAS201853}. In another example, Mbuwir et. al. manage a battery directly using batch Q-learning \citep{en10111846}. 
  
    However, there are few works where the RL controller is a price setter for demand response in an office. To the best of our knowledge, no one has used MAML to improve the learning of the RL controller over a vanilla controller. We are proposing this agent for use in a novel experiment with demand response within a social game.

 \section{Methods}\label{sec:methods}
 
 
 We adapt MAML to the problem of optimizing a price-serving agent for energy demand response. We will briefly explain the baseline PPO architecture. We will then describe the simulation environment we test this in. Finally, we will then explain the MAML implementation and how we tested it in the environment. 
 
  \begin{figure}
\centerline{\includegraphics[width=0.9\linewidth]{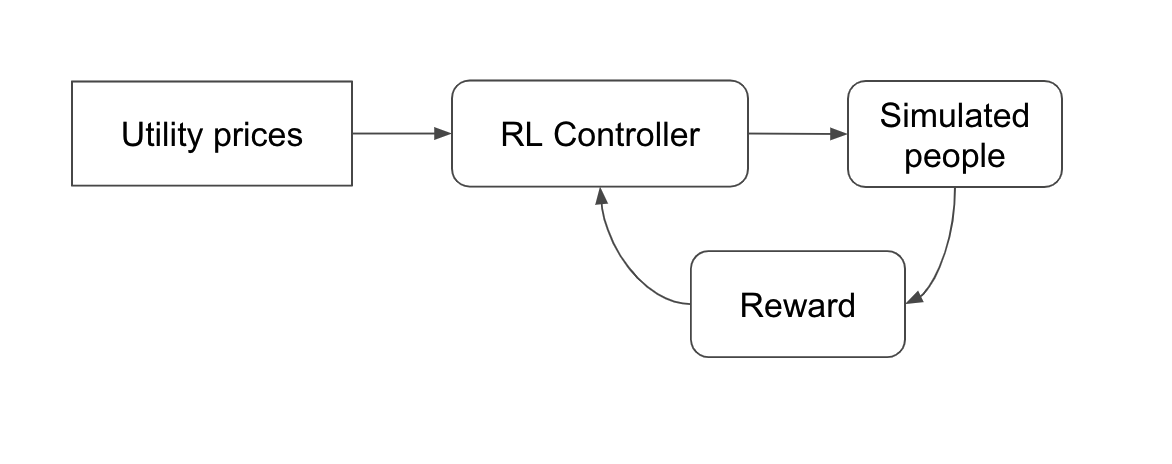}}
\caption{Reinforcement Learning Control Flow} \label{fig:RLcontroller}
\end{figure}

Can an RL agent preemptively estimate the most effective demand response price using historical data and implicitly predicting causal factors? Furthermore, can we pretrain an agent in simulation that can quickly adapt its predictions to real-world data? We employ Proximal Policy Optimization (PPO) and Model-Agnostic Meta-Learning (MAML) to train an agent in OfficeLearn to optimize our policy network to adapt quickly to new environments. We use the RLLib \cite{liang2018rllib} implementation of PPO and MAML, and the default RLLib neural network architecture of two hidden layers with 256 units each and $\tanh$ activations.  The reward for the price-setting agent is $-\log(d^tg) - \lambda * \mathbbm{1}(d^tg - \hat{d})$, where $d$ is the demand of the person it studies, $g$ is the grid pricing, $\hat{d}$ is some minimal baseline energy demand that the grid must meet, and $\lambda * \mathbbm{1}(d^tg - \hat{d})$ is a regularizing penalty that is applied if the simulated energy grid does not meet a mean baseline level of energy demand. The penalty is designed to avoid the locally optimal solution of the agent driving down prices indiscriminately without regard for energy supply. For other implementation choices, please see our Github\footnote{Please find our Github repo at the following link: \url{https://github.com/Aphoh/temp_tc/tree/austin_maml}} and RLLib.
 
\subsection{Environment}
 
We summarize an OpenAI gym environment built to simulate demand response in office buildings \cite{spangherofficelearn}.  Each step in the environment is a day, where the agent proposes prices to office workers. Based on the agent's prices, the office workers modify their energy consumption behaviors in order to achieve the lowest cost of energy possible, which the controller then uses to assign points. Each simulated person has a deterministic response to the points that are offered to them. Notably, we employ several different models of simulated response, with two levels of complexity: ``Deterministic Function'' and ``Curtail and Shift''. Their descriptions are listed below:

\subsubsection{``Deterministic Function'' Person}

We include three types of deterministic response within one type of agent, with the option of specifying a mixed office composed of all three types. 

A "Deterministic Function" Person with \textbf{linear response} decreases their energy consumption linearly below an average historical energy consumption baseline. Therefore, if $b_t$ is the historical baseline at time $t$ and $p_t$ are the points given and $m$ is a simulation set points multiplier, the energy demand $d$ at time $t$ is $ d_t = b_t - p_t * m$, clipped at ceiling and floor values $d_{min}$ and $d_{max}$, which are 5\% and 95\% of a historical energy distribution (i.e. a distribution made from two months worth of historical data from \citep{spangher2019visualization}.) 
A "Deterministic Function" Person with \textbf{sinusoidal response} is one who responds to points towards the middle of the distribution and not well to low or high points. Therefore, the energy demand $d$ at time $t$ is $ d_t = b_t - \sin{p_t} * m$, clipped at $d_{min}$ and $d_{max}$.

A "Deterministic Function" Person with \textbf{threshold exponential response}, we define an office worker who does not respond until points pass a threshold, at which point they respond exponentially. Therefore, the energy demand $d$ is $d_t = b_t - (\exp{p_t} * ( p_t > 5))$ , clipped at $d_{min}$ and $d_{max}$. 
 
\subsubsection{``Curtail And Shift Office Worker''}

Office workers need to consume electricity to do their work, and may not be able to curtail their load below a minimum threshold, e.g. the minimum power needed to run a PC. They may have the ability to shift their load over a definite time interval, e.g. choosing to charge their laptops ahead of time or at a later time. We model a response function that exhibits both of these behaviors. We can model the aggregate load of a person ($b_t$) as a combination of fixed demand ($b^{fixed}_t$), curtailable demand ($b^{curtail}_t$), and shiftable demand ($b^{shift}_t$), i.e., $b_t = b^{fixed}_t + b^{curtail}_t + b^{shift}_t$. All of the curtailable demand is curtailed for the $T_{curtail}$ hours (set to $3$ hours in practice) with the highest points, and for every hour $t$ the shiftable demand is shifted to the hour within $[t - T_{shift}, t+T_{shift}]$ with the lowest energy price. For example, such an office worker may need to charge their appliances for a total of 1000 Wh throughout the day, 300 of which are for printing documents that could be curtailed, 300 of which are for presenting at a meeting whose time can be shifted, and 400 of which are the minimum energy requirement to run a computer. Upon receiving a price signal with high prices from 11am-2pm, this simulated worker would curtail their printing away from the 3 hours with the highest energy price, and schedule their meeting at the hour within $[t - T_{shift}, t + T_{shift}]$ with the lowest energy price. This would allow them to decrease their energy usage.

\subsection{Simulated Experiment}
 To test our hypothesis that MAML + PPO will enable faster adaptation to unfamiliar environments like a real-world Social Game, we trained MAML on several models of simulated person response. We then evaluated how quickly PPO, starting from the MAML weight initialization, can learn in an OfficeLearn environment with different models of simulated person response. The training environments had randomized "Deterministic Function" response types and multipliers for how many "points" simulated humans received for reducing energy usage. Though the training environments used to train MAML had randomized parameters, the validation environments were kept constant to ensure fairness. To ensure an accurate representation of each network's capabilities, we averaged the results from 5 different test trials and report the mean and standard error for each test. MAML is trained with an ADAM optimizer with learning rate 0.0001, 0.9 $\beta_1$, and 0.999 $\beta_2$, where $\beta_1$ refers to the first moment and $\beta_2$ refers to the second.  MAML sampled approximately eight PPO training trajectories in (parallel) training environments at a time between (sequential) meta-update steps. We trained PPO for 5 steps at a time in MAML's inner adaptation phase and trained MAML once per trajectory sampling. We trained MAML for up to 200 iterations. For evaluation, we trained PPO for 100 days (i.e. PPO's iterations) from the MAML-learned weight initialization and from a random initialization and compared the two. PPO (in both MAML+PPO and PPO) is trained with the clipped surrogate loss with a clipping parameter of 0.3 and a SGD optimizer with learning rate 0.01. The action and value estimators shared layers in the neural network for our implementation, and the value loss had a weighting of 0.5. We present three different experiments: 
 \subsubsection{Adaptation to "Curtail and Shift" Response}
We train MAML on a distribution of environments with three different models of human response to price: "Deterministic Function" with all responses and see if it can adapt to a completely different model that we believe may be more representative of real-world behavior: "Curtail and Shift", where testing was done over 5 different test trials of each.

 \subsubsection{Adaptation to Threshold Exponential Response}
 
In this experiment, we train MAML on a distribution of environments with two different models of human response to price instead of three: "Deterministic Function" with linear or sinusoidal responses. Each variant environment, along with other parameters, are chosen at random when MAML samples from the task distribution during training. We test this trained model's ability to adapt to an environment with a different model of human response: "Deterministic Function"Person with threshold exponential response, with all 5 validation runs of MAML+PPO and PPO.

\subsubsection{MAML Training Behavior}

We observe the performance of MAML + PPO in the \textit{Adaptation to Curtail and Shift Response} task after several different numbers of MAML training iterations. When training MAML, we saved checkpoints of the meta-optimized weight initialization at 50, 100, 150, and 200 MAML iterations. Then, we observed the performance of PPO starting from each of those warm starts. 

 \section{Results}\label{sec:results}
 We will now describe the results obtained from using MAML. 
 
 \subsection{MAML Results (Compared to baseline)}
\begin{figure}
\centerline{\includegraphics[width=0.95\linewidth]{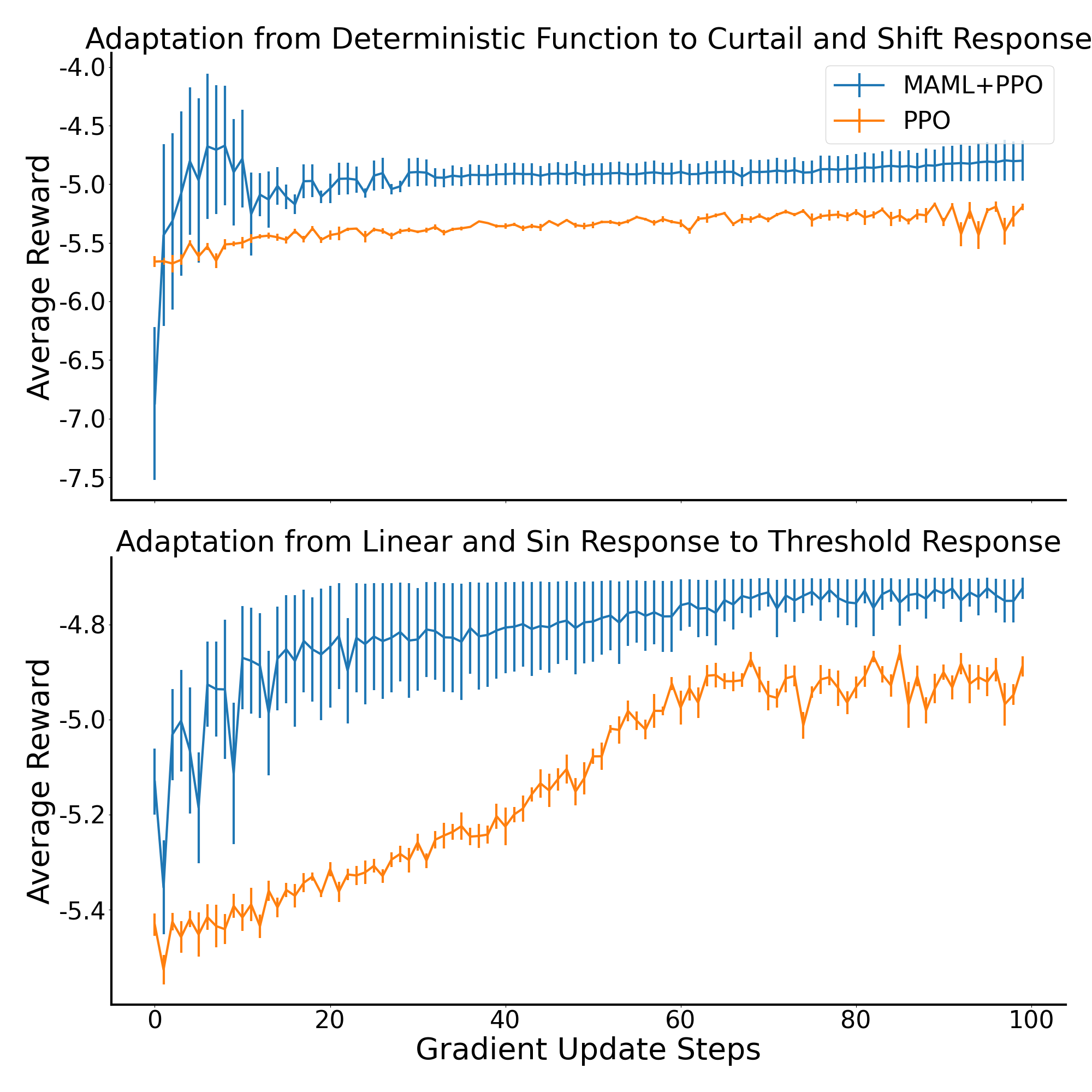}}
\caption{MAML+PPO Results}\label{fig:MAMLResults}
\medskip
\small
\begin{flushleft}
\textbf{A.} Performance of MAML+PPO in training on a simpler model of human behavior, "Deterministic Function", and adapting to a more complex model, "Curtail and Shift" in comparison to PPO. \textbf{B.} Performance of MAML+PPO in a lateral shift in complexity, i.e. adapting to a different model of human behavior, "Deterministic Threshold Exponential" in comparison to PPO.
\end{flushleft}
\end{figure}
As can be seen in Fig. \ref{fig:MAMLResults}.A and B, MAML + PPO significantly outperforms baseline PPO from the start to the finish of training. In both adaptation experiments, MAML+PPO converges to a solution with higher reward, and thus lower simulated energy cost than PPO. Note that PPO appears to have plateaued with constant reward. Our addition of MAML decreases total energy cost by about 40\% in both environments compared to PPO by the end of training on average. It takes fewer than 2 simulated days (i.e. iterations) worth of training data for MAML+PPO to outperform PPO after training for 100 simulated days, which suggests that the use of MAML with PPO to warm start learning from our simulation was successful. Using MAML allowed us to find a weight initialization that generalized to more complex tasks, thus decreasing the amount of data necessary to get the controller to a certain level of performance. 

\begin{figure}
\centerline{\includegraphics[width=0.95\linewidth]{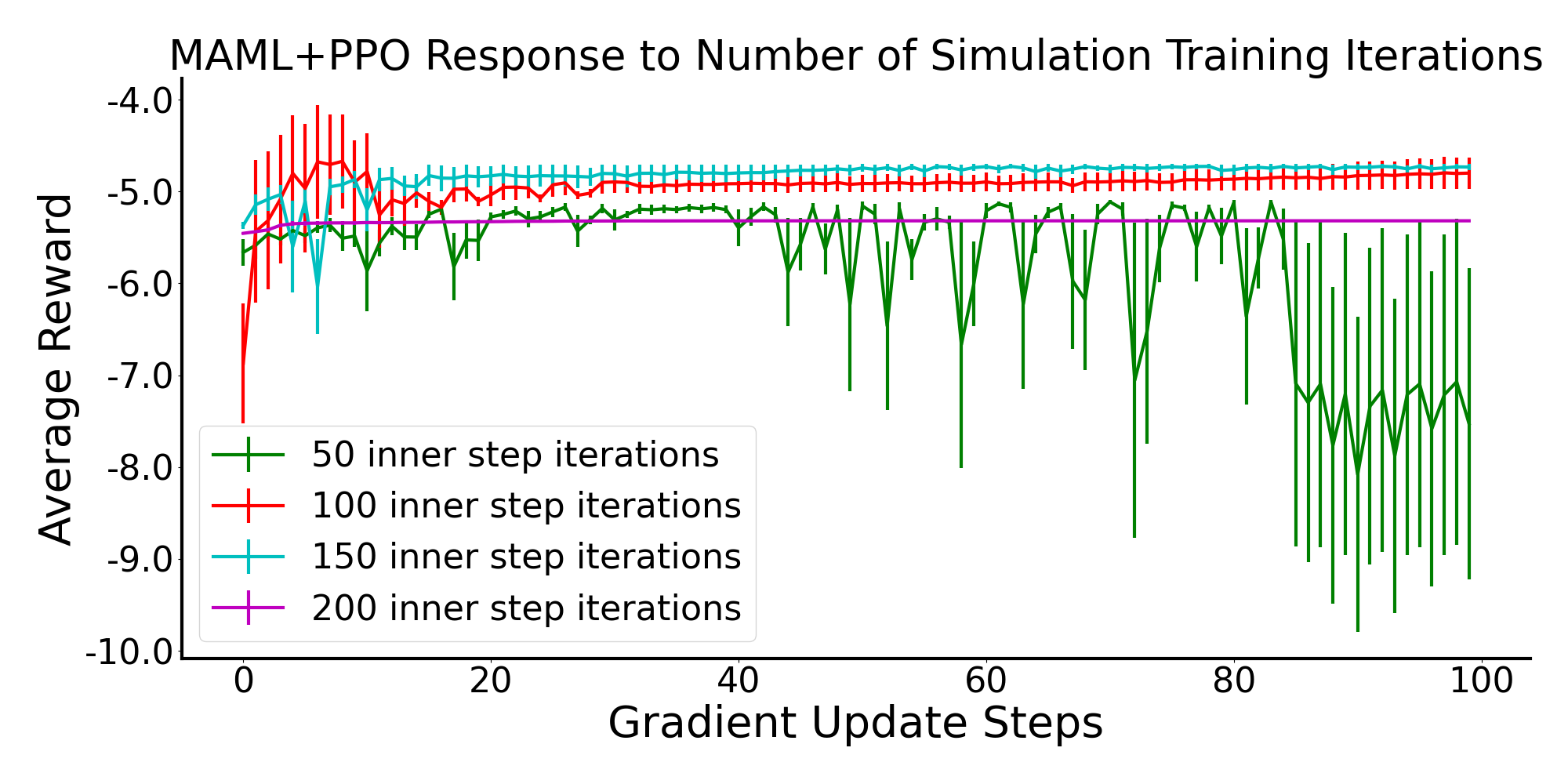}}
\caption{MAML Reward vs Update Steps}\label{fig:MAMLAblation}
\medskip
\small
\begin{flushleft}
Performance of MAML+PPO on the "Curtail and Shift" adaptation task after different number of iterations of training MAML in simulation.
\end{flushleft}
\end{figure}

As we are interested in both the reward achieved by each algorithm and how quickly the reward is achieved, we chose to plot the training graphs of the validation runs of each checkpoint in Fig. \ref{fig:MAMLAblation} instead of simply plotting the total reward of each validation run. We observe that performance appears to increase with number of MAML training steps up until 200 iterations, where MAML appears to have overfit to a local minimum.  Training MAML for too few steps, for example only 50 iterations, appears to increase instability in the PPO training on the MAML weight initialization.
 
 \section{Discussion}\label{sec:Discussion}
  
Meta-learning and K-shot learning are fields of machine learning focused on applying knowledge obtained from one domain to another. This makes techniques like MAML especially helpful in taking advantage of simulations, where data is plentiful, to increase performance in the real world, where data is costly. Whereas other RL tasks like robotic grasping can use highly accurate physics simulations to train a model and successfully employ the same model with minimal changes to real world equivalent tasks, the human response to changes in electricity price is not as well studied as physics. The models of human behavior we employ in simulations are often simplistic and likely inaccurate with regard to real human behaviour. However, the novelty of our experiment is in showing how MAML might be applied to transfer knowledge acquired from training in simulation to accelerating adaptation in the real world, by demonstrating that MAML allows fast adaptation to different models of human responses to energy price. From our results, we are hopeful that MAML+PPO can be applied in real world price controller RL as well. By training MAML+PPO in an environment different from the test environment, we were able to affirm that the technique can be used to adapt training in simulation to a different task in simulation. The jump in complexity between the "Deterministic Function" response the model was trained on and the "Curtail and Shift" response shows that MAML+PPO can be used to adapt to tasks that are outside of its training task distribution and are more complex. We believe that this ability for MAML to close gaps in complexity has a high probability of enabling a MAML optimized model to work as a real world RL price controller. 

\subsection{Future Research}

We hope to integrate MAML with other enhancements that we are testing in the learning controller: namely, a Surprise-Minimizing RL reward regularizer, detailed in a different submission to this same conference, and a multi-agent architecture of TD3. We would like to test this in simulation.  

More practically, we will use this enhancement directly in the rollout of our agent for our office-scale demand response experiment. We present in this work a training on a distribution of tasks defined by simpler people and tested on more complex people, but we will essentially train the experiment controller on the superset of the presented training and validation simulations, and test it on real people. This will allow us to confirm our hypothesis that MAML allows for faster adaptation from simulation to the real world when training an RL price controller.

\subsection{Acknowledgements}


We gratefully acknowledge the input of Manan Khattar, William Arnold, and Tarang Srivistava as working group collaborators who met weekly on their own research thrusts and provided advice (and at times a listening ear.) We give thanks to Andreea Bobu and Peter Hendrickson who are graduate students that gave guidance during this process. Finally, we would like to thank Chelsea Finn for her initial work on MAML and Pieter Abbeel and Sergey Levine for advice throughout the process. 

\bibliographystyle{ACM-Reference-Format}
\bibliography{sample-base}

\end{document}